\documentclass{article}
\usepackage{preprint,times}

\usepackage{amsmath,amsfonts,bm}

\def\eqref#1{equation~\ref{#1}}

\def\1{\bm{1}}

\DeclareMathAlphabet{\mathsfit}{\encodingdefault}{\sfdefault}{m}{sl}
\SetMathAlphabet{\mathsfit}{bold}{\encodingdefault}{\sfdefault}{bx}{n}

\usepackage{hyperref}
\hypersetup{hidelinks}
\usepackage{url}
\usepackage{graphicx}
\usepackage{float}
\usepackage{tikz}
\usepackage{booktabs}

\title{Rollout-Marginal Distillation for Long-Horizon Autoregressive Video Generation}

\author{\textbf{Chenjian Gao}$^{1,2*}$,\quad
\textbf{Zhihao Hu}$^{2*}$,\quad
\textbf{Jianqi Ma}$^{2}$,\\[3pt]
\textbf{Jun Zhang}$^{2\dagger}$,\quad
\textbf{Weidong Zhang}$^{2}$,\quad
\textbf{Tianfan Xue}$^{1\dagger}$\\[8pt]
$^{1}$MMLab, The Chinese University of Hong Kong\quad $^{2}$Tencent AIPD\\[3pt]
$^{*}$Equal contribution\qquad $^{\dagger}$Corresponding authors
}

\date{}

\hypersetup{
  pdftitle={Rollout-Marginal Distillation for Long-Horizon Autoregressive Video Generation},
  pdfauthor={Chenjian Gao, Zhihao Hu, Jianqi Ma, Jun Zhang, Weidong Zhang, Tianfan Xue}
}

\begin{document}

\maketitle

\begin{abstract}
Autoregressive (AR) video diffusion enables low-latency, streamable video generation, but prediction errors often accumulate over long rollouts. Training the generator on its own rollouts exposes it to these imperfect histories. However, existing video-level distribution matching distillation (DMD) scores the whole rollout jointly. Because a chunk is evaluated together with its past and future, its correction can favor matching artifacts in the surrounding context merely to preserve temporal consistency. To provide a clearer visual-quality signal, we introduce Rollout-Marginal Distillation (RMD). RMD retains the generated history for AR prediction but scores each chunk independently against a chunk teacher, ensuring its quality correction is not compromised by an imperfect temporal context. To compensate for the lack of temporal context in independent chunk scoring, RMD subsequently applies video-level DMD to restore temporal coherence. Extensive experiments demonstrate that RMD maintains high visual quality far beyond its training horizon and outperforms video-level DMD baselines. Code and video results are available at \url{https://cjeen.github.io/RMD/}.
\end{abstract}

\begin{figure}[H]
\centering
\includegraphics[width=0.90\linewidth]{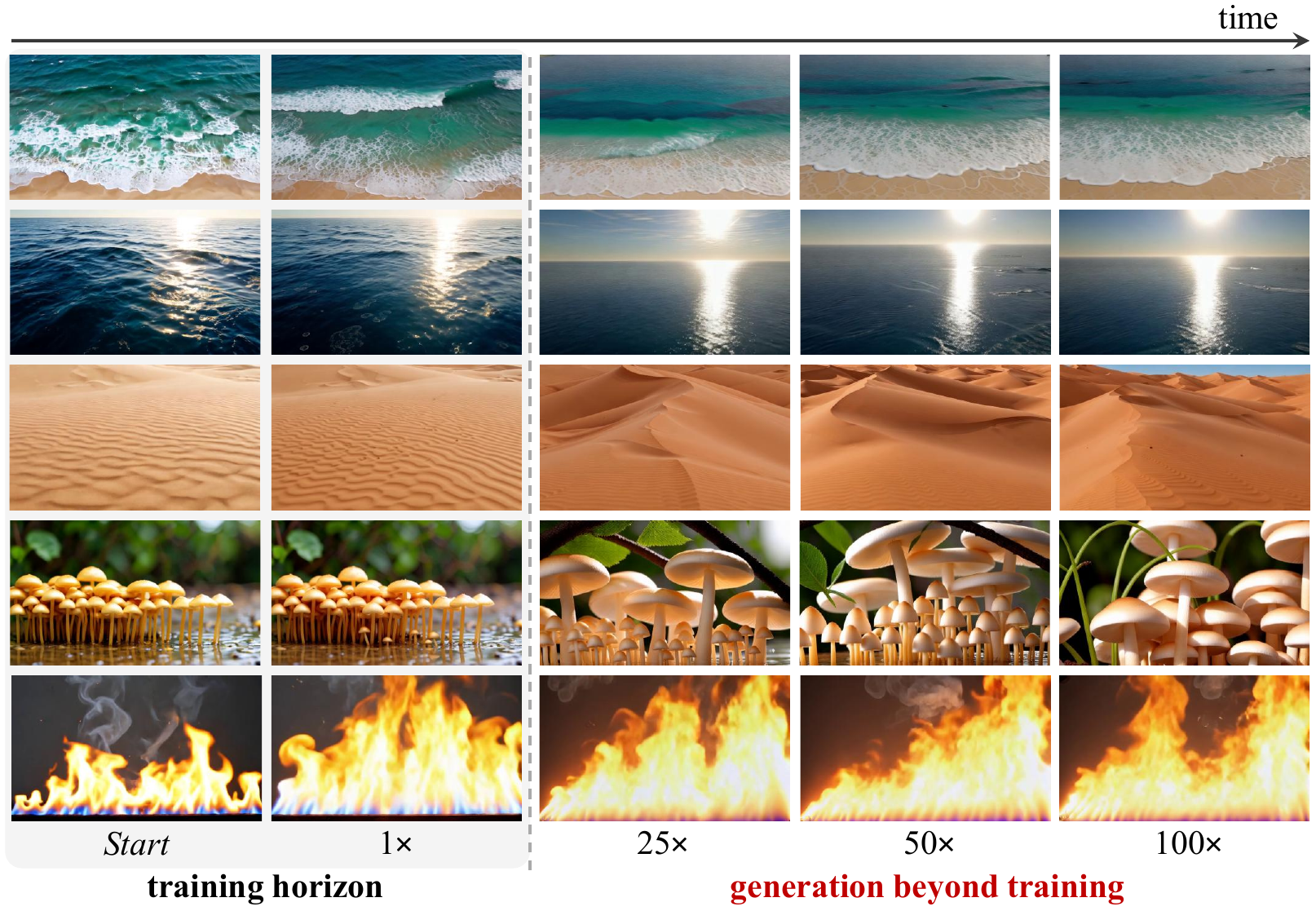}
\caption{\textbf{Long-horizon autoregressive video generation.} Trained on 5-second
rollouts, RMD maintains high visual quality over 500-second rollouts
($100\times$ the training horizon). Notably, generation relies solely on a fully sliding
context window, without fixed anchors,
additional explicit memory mechanisms, or modifications to the generator architecture.}
\label{fig:rmd}
\end{figure}

\clearpage
\section{Introduction}

\begin{figure}[t]
    \centering
    \includegraphics[width=\linewidth]{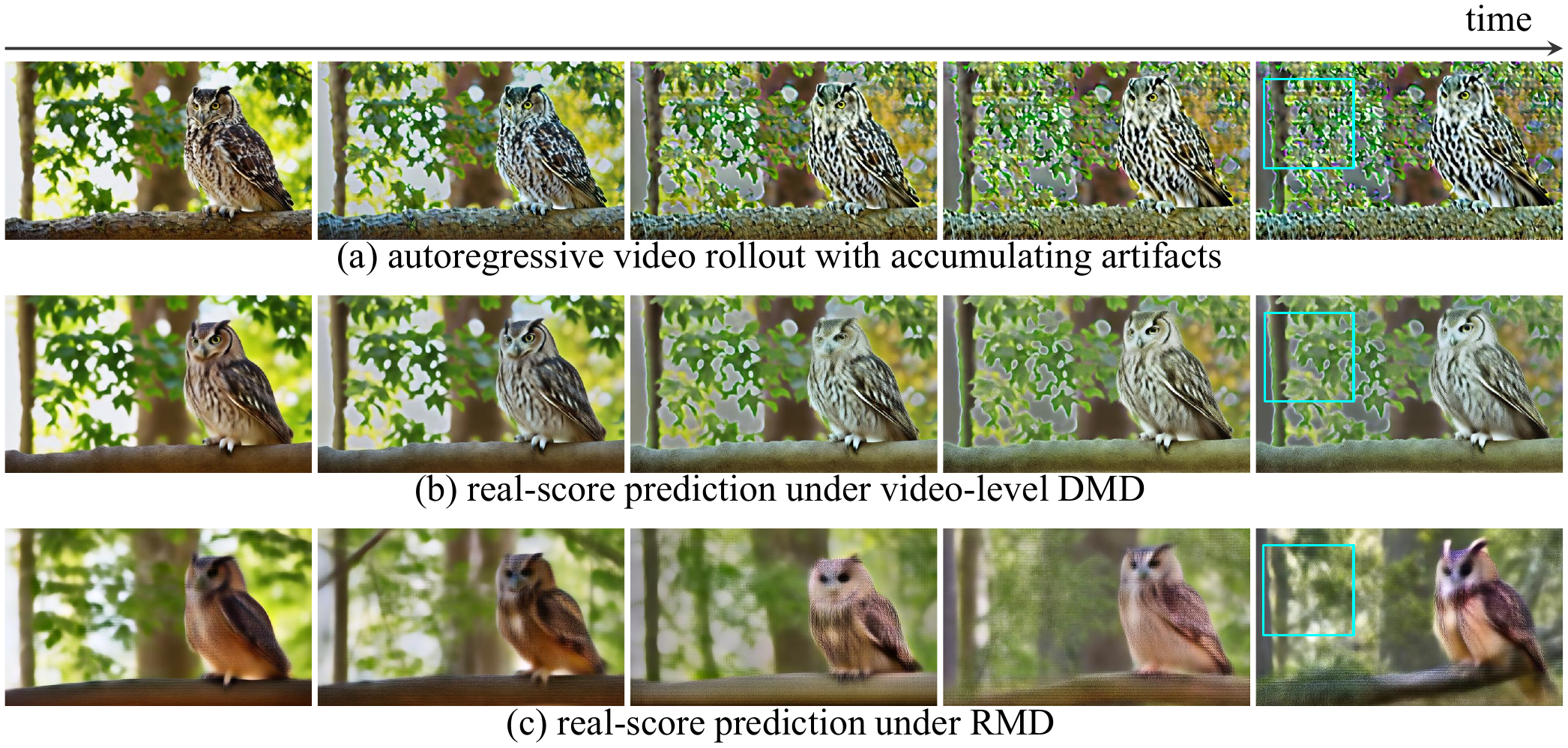}
    \caption{\textbf{Comparison of real-score predictions under video-level DMD and RMD.} (a) The autoregressive rollout accumulates visual artifacts. (b) Video-level DMD scores the sequence jointly, causing the real-score prediction to retain these historical artifacts. (c) RMD scores the current chunk independently, providing a clean target regardless of the degraded context.}
    \label{fig:supervision_comparison}
\end{figure}

Due to high computational and memory costs, video diffusion models are typically trained on short clips~\citep{wan2025wan,huang2025selfforcing}. However, many real-world applications demand much longer sequences, including autonomous-driving simulation, interactive and embodied world models, and physics-grounded visual simulation~\citep{zhang2025epona,bruce2024genie,liu2024physgen}. Autoregressive (AR) models naturally support this by extending video generation chunk by chunk~\citep{yin2025causvid,huang2025selfforcing}. Specifically, the diffusion process for each new chunk conditions on previously generated frames. As the video grows, a sliding context window ensures the model only attends to the most recent history, keeping the generation cost bounded. However, this sequential process introduces an inherent challenge: each generated chunk inevitably contains minor prediction errors, which then become the context for future steps. Consequently, errors barely noticeable in a short sequence can compound and severely degrade visual quality over a long rollout.

To mitigate error accumulation, self-rollout training exposes the generator to its own imperfect predictions~\citep{huang2025selfforcing,zhu2026causal}. Since a generated video rarely matches real videos perfectly, no paired ground-truth continuation exists to supervise the next steps. To overcome this lack of paired data, distribution matching distillation (DMD) is often employed~\citep{yin2023dmd,yin2025causvid}. Instead of requiring exact continuations, a bidirectional diffusion teacher evaluates the entire sequence jointly. It provides corrective feedback by matching a fake score tracking the student's output against a real score representing the teacher's high-quality prior.

However, DMD-trained models still struggle to generate very long sequences. We attribute this to a bias introduced by the video teacher's joint supervision. As visual artifacts accumulate over a long autoregressive rollout (e.g., the severe texture and color corruption in the background foliage in Figure~\ref{fig:supervision_comparison}(a)), the teacher must evaluate the new chunk alongside this flawed history. Because the teacher cannot modify past frames, enforcing a clean, natural background in the current chunk would cause an abrupt visual shift. Consequently, to preserve temporal consistency, the video-level teacher often retains these accumulated errors, which manifest as non-physical ghosting artifacts, in its real-score prediction (Figure~\ref{fig:supervision_comparison}(b)). The temporal consistency constraint thus fundamentally limits the teacher's ability to correct local errors.

We break this coupling with Rollout-Marginal Distillation (RMD). The generator still uses its history to produce continuations, but the teacher evaluates the generated chunk independently, without checking the history or future chunks. As shown in Figure~\ref{fig:supervision_comparison}(c), context-free scoring ensures the teacher provides a clean, high-fidelity real score despite the degraded history. RMD achieves this by matching the student's chunk marginal distribution to a teacher-defined chunk distribution. To provide this supervision, we prepare the RMD teacher by adapting a video diffusion model on individual chunks extracted from videos generated by the original bidirectional model. However, because independent chunk supervision does not enforce cross-chunk relationships, generated videos might exhibit motion jitter between chunks. To resolve this, we subsequently refine the generator with a video-level DMD stage, explicitly smoothing these transitions to promote temporal coherence.

Extensive experiments demonstrate that RMD achieves significantly higher visual quality than video-level DMD baselines beyond the training horizon. Our quantitative evaluation covers rollouts of approximately 60 seconds ($12\times$ the 5-second training horizon). In addition, Figure~\ref{fig:rmd} provides a qualitative example extending to 500 seconds ($100\times$ the training horizon). Generation relies entirely on a fully sliding context window, without fixed anchors such as frame sinks~\citep{yang2026longlive,li2026rollingsink}, additional explicit memory mechanisms, or changes to the generator architecture.

\section{Related Work}

\paragraph{Few-step autoregressive video generation.}
Causal video diffusion generates frames or blocks sequentially with KV caching for
low-latency streaming. CausVid converts a pretrained bidirectional video diffusion model into a
causal generator and extends DMD to few-step video generation~\citep{yin2025causvid}. Self-Forcing
further performs autoregressive rollout during training, feeding generated outputs back as context
and supervising the sequence with a holistic video objective~\citep{huang2025selfforcing}.
Causal Forcing instead studies initialization and uses an autoregressive teacher for ODE
distillation before applying Self-Forcing DMD~\citep{zhu2026causal}. Causal-rCM similarly combines
teacher-forced consistency training with self-forced DMD refinement~\citep{zheng2026causalrcm}.
Solaris introduces \textit{Checkpointed Self Forcing}, which decouples serial self-rollout from parallel
differentiable replay to reduce training memory~\citep{savva2026solaris}. Together, these methods
establish causal students and on-policy training pipelines. We adopt the Solaris replay
mechanism directly and focus RMD on the distribution matched during on-policy DMD.

\paragraph{Distribution matching distillation.}
DMD trains a few-step generator by minimizing a reverse KL divergence to the distribution of a
pretrained diffusion model, using the difference between real and fake scores as generator
supervision~\citep{yin2023dmd}. DMD2 removes the costly regression objective and improves stability
through a two-time-scale update rule, adversarial learning, and multi-step training~\citep{yin2024dmd2}.
Subsequent work has combined distribution matching with adversarial learning~\citep{lu2025adm}, replaced KL-based
matching with an optimal-transport objective~\citep{wang2026vdot}, or coupled reverse-KL refinement
with score-regularized consistency training~\citep{zheng2026rcm}. These developments primarily
modify the divergence, regularization, or optimization of distribution matching. By contrast, RMD
retains the DMD score-difference update but changes its output space from a joint rollout to the
chunk marginal induced by autoregressive sampling.

\paragraph{Long rollouts and causal supervision.}
LongLive improves extended AR generation through streaming long-video tuning, a finite attention
window, and a persistent frame sink~\citep{yang2026longlive}. Rolling Sink studies cache
maintenance beyond the training horizon and introduces a training-free cache-maintenance
strategy~\citep{li2026rollingsink}. More directly related to training-time supervision, OPSD-V
post-trains few-step AR generators on their inference-time trajectories while supplying the teacher
with a cleaner cache that can incorporate real long-video context~\citep{liu2026opsdv}. Most closely related to our
analysis, Context-Matched Distillation (CMD) observes that a bidirectional full-clip teacher can use
future information unavailable to a causal student~\citep{bandyopadhyay2026cmd}. CMD addresses
this issue with a causal teacher and prefix scoring under the student's realized rollout context.
RMD takes a different route. It does not construct a teacher conditional for each self-rollout
history. Instead, it uses a bidirectional diffusion prior as a chunk-distribution target, scores the
rollout chunks independently, and subsequently restores joint temporal structure through video-level
refinement. It therefore requires neither paired continuations for generated histories nor real
long-video context.

\section{Rollout-Marginal Distillation}
\label{sec:rmd}

In this section, we present Rollout-Marginal Distillation (RMD), an approach designed to disentangle visual quality supervision from temporal context in autoregressive video generation. We first review video-level distribution matching distillation (DMD) and formalize how its joint scoring mechanism inherently couples the current chunk's gradient to past and future frames (Section \ref{sec:revisiting}). To break this coupling, we introduce our rollout-marginal objective, which independently penalizes unrealistic chunks without conditioning on the imperfect context (Section \ref{sec:objective}). We then detail the training procedure in practice (Section \ref{sec:instantiation}). Finally, because independent scoring ignores cross-chunk relationships, we describe a subsequent video-level refinement stage that explicitly restores temporal coherence (Section \ref{sec:refinement}).

\subsection{Revisiting Video-Level Distribution Matching}
\label{sec:revisiting}

To understand the necessity of rollout-marginal supervision, we first examine how video-level DMD improves a causal video generator and why its joint scoring mechanism introduces a critical flaw. 

Given a text prompt \(c\), let the causal generator \(G_\theta\) produce a latent rollout \(\mathbf{x}=(x_1,\ldots,x_T)\), where each \(x_i\) is a chunk. The distribution of the generated video \(\mathbf{x}\) can be naturally factorized as:
\begin{equation}
    q_\theta(\mathbf{x}\mid c)=\prod_{i=1}^{T}q_\theta(x_i\mid x_{<i},c).
    \label{eq:causal_factorization}
\end{equation}
Following the self-rollout framework~\citep{huang2025selfforcing,zhu2026causal}, the generator conditions each continuation on its own generated history \(x_{<i}\), exposing it to the prediction errors it will encounter at inference. Because there are no ground-truth continuations paired with these imperfect generated histories, video-level distribution matching distillation (DMD) is employed to supervise the entire student rollout jointly~\citep{yin2023dmd,yin2024dmd2,yin2025causvid}. 

At noise level \(\tau\), let \(\mathbf{x}_\tau=\alpha_\tau\mathbf{x}+\sigma_\tau\boldsymbol{\epsilon}\) denote a noised rollout. The video-level objective minimizes the reverse KL divergence between the noised student video distribution \(q_{\theta,\tau}\) and a pretrained teacher video distribution \(p_\tau\):
\begin{equation}
    \mathcal{L}_{\mathrm{video}}(\theta)
    =\mathbb{E}_{c,\tau}\!\Bigl[
        D_{\mathrm{KL}}\!\bigl(
            q_{\theta,\tau}
            \,\Vert\,
            p_\tau
        \bigr)
    \Bigr].
    \label{eq:video_dmd_objective}
\end{equation}

The generator is updated using the difference between two diffusion scores:
\begin{equation}
    \nabla_\theta\mathcal{L}_{\mathrm{video}}
    \approx\mathbb{E}\!\Bigl[
        w(\tau)J_\theta(\mathbf{x})^\top
        \bigl(s_{\mathrm{fake}}(\mathbf{x}_\tau,\tau,c)
        -s_{\mathrm{real}}(\mathbf{x}_\tau,\tau,c)\bigr)
    \Bigr],
    \label{eq:video_update}
\end{equation}
where \(J_\theta(\mathbf{x})=\partial\mathbf{x}/\partial\theta\) is the generator Jacobian, \(w(\tau)\) is a noise-level weight, \(s_{\mathrm{real}}\) is a frozen teacher score representing the target distribution, and \(s_{\mathrm{fake}}\) is an online score model trained to track the student’s distribution.

\paragraph{The Coupling Dilemma.} 
Equation \ref{eq:video_update} reveals a fundamental issue in applying video-level DMD to autoregressive rollouts. Both the real and fake score networks (\(s_{\mathrm{real}}\) and \(s_{\mathrm{fake}}\)) take the entire sequence \(\mathbf{x}_\tau\) as input. Consequently, the correction signal assigned to any specific chunk \(x_i\) is coupled with its generated past (\(x_{<i}\)) and future (\(x_{>i}\)). If an artifact appears in the generated history, a joint score evaluation will likely penalize a sudden correction to natural colors in the current chunk, as such a transition breaks the temporal consistency of the sequence. The model is thereby forced to compromise local visual realism to accommodate an imperfect temporal context. To resolve this, we must disentangle the quality supervision of \(x_i\) from the rest of the rollout.

\subsection{Rollout-Marginal Objective}
\label{sec:objective}

Video-level DMD scores the entire sequence, tying each chunk's quality correction to its compatibility with the surrounding content. RMD breaks this dependency by matching chunks from the causal rollouts independently, while the generator still uses its history for prediction.

We model the initial and continuation chunks separately to account for their distinct latent statistics. Specifically, under the video VAE, the first latent frame encodes only the initial image without temporal compression~\citep{wan2025wan,huang2025selfforcing}. This gives it completely different statistical properties compared to subsequent frames. Let \(q_{\theta,1}(x_1\mid c)\) denote the initial chunk distribution. To define the continuation \emph{rollout marginal}, we pool the chunks \(x_2, \ldots, x_T\) across the student's causal rollouts for prompt \(c\):
\begin{equation}
    q^{(T)}_{\theta,\mathrm{chunk}}(x\mid c)
    =\frac{1}{T-1}\sum_{i=2}^{T}
    \mathbb{E}_{x_{<i}\sim q_\theta(x_{<i}\mid c)}
    \!\left[q_\theta(x\mid x_{<i},c)\right].
    \label{eq:rollout_marginal}
\end{equation}
This distribution averages over both the generated histories and the continuation positions. Sampling from it requires the causal rollout, but evaluating a chunk's realism requires neither its history nor its rollout position.

At noise level \(\tau\), let \(p_{1,\tau}\) and \(p_{\mathrm{chunk},\tau}\) denote the teacher-defined target distributions for the initial frame and continuation chunks, respectively. Weighting these two groups by their frequency in the rollout yields the RMD objective:
\begin{equation}
    \mathcal{L}_{\mathrm{RMD}}(\theta)
    =\mathbb{E}_{c,\tau}\!\left[
        \frac{1}{T}D_{\mathrm{KL}}\bigl(q_{\theta,1,\tau}\,\Vert\, p_{1,\tau}\bigr)
        + \frac{T-1}{T}D_{\mathrm{KL}}\bigl(q_{\theta,\mathrm{chunk},\tau}^{(T)}\,\Vert\, p_{\mathrm{chunk},\tau}\bigr)
    \right].
    \label{eq:rmd_objective}
\end{equation}

Crucially, the generation process remains causal, meaning each sample \(x_i\) is still produced conditionally from its actual on-policy history \(x_{<i}\). However, the supervision applied to each chunk is entirely context-free. This objective simply penalizes generated chunks that are unlikely under the teacher's chunk distribution. By doing so, it explicitly isolates visual quality supervision from the imperfect generated history. Instead of forcing the model to repeat past artifacts to maintain temporal consistency, this marginal matching ensures the teacher guides the current chunk toward high-quality targets, regardless of the flaws in the preceding frames.

\begin{figure}[t]
    \centering
    \includegraphics[width=\linewidth]{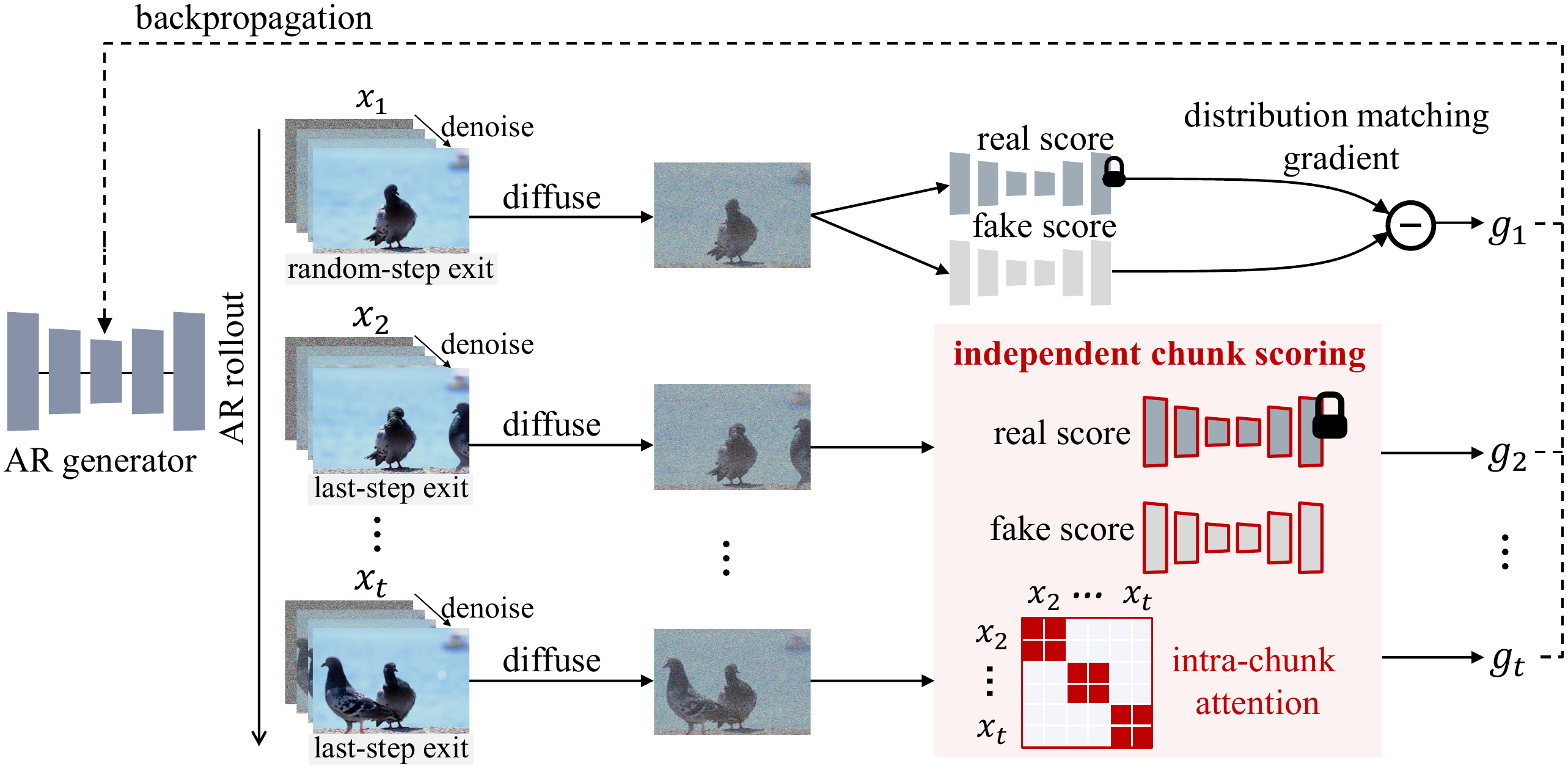}
    \caption{\textbf{Overview of Rollout-Marginal Training.} The AR generator produces a causal rollout, sampling the initial chunk ($x_1$) at a random-step exit and continuation chunks ($x_2, \dots, x_T$) at the last-step exit. Adapted score networks evaluate them independently without temporal context. The resulting gradients ($g_1, \dots, g_T$) are backpropagated via differentiable causal replay.}
    \label{fig:method}
\end{figure}

\subsection{Rollout-Marginal Training}
\label{sec:training}
\label{sec:instantiation}

To realize this decoupled supervision in practice, our training pipeline (Figure \ref{fig:method}) optimizes the rollout-marginal objective by first generating the causal rollout, then evaluating the initial and continuation chunks against their distinct target distributions, and finally backpropagating the isolated gradients to the generator.

We initialize the AR generators from Causal Forcing's causal ODE checkpoints~\citep{zhu2026causal}. Following the self-forcing framework~\citep{huang2025selfforcing}, at each training iteration, we generate a causal rollout of chunks \(x_1, \ldots, x_T\), sampling each chunk in 4 denoising steps. To retain the temporal dynamics prior in the pretrained model, we carefully choose which denoising step to supervise for each chunk~\citep{wu2026dpdmd}. Early steps in the sampling trajectory rely heavily on historical context to establish global motion, and applying independent supervision at these timesteps would disrupt the temporal consistency. Therefore, we adopt an asymmetric supervision strategy: for the initial chunk \(x_1\), we compute the loss at a randomly selected intermediate exit step along the trajectory, whereas for the continuation chunks \(x_2, \ldots, x_T\), we apply the loss only at the final, lowest-noise step of the rollout where the network refines local visual details.

To score the sampled chunks, we use separate teacher targets for the initial frame and continuations. As discussed in Section~\ref{sec:objective}, Wan's causal VAE gives them different latent statistics. An unmodified video teacher would treat an isolated continuation chunk as the start of a video and expect an uncompressed first frame. We therefore adapt the Wan 14B video teacher~\citep{wan2025wan} with LoRA~\citep{hu2022lora} on continuation chunks extracted from videos generated by the original model. The adapted teacher supplies the continuation target \(p_{\mathrm{chunk},\tau}\) in Equation~\ref{eq:rmd_objective}, allowing each continuation to be scored without its generated history.

For each \(x_i\), we add scoring noise and obtain clean predictions from its teacher and fake-score networks on the same noisy input. Let \(D_{\mathrm{real},i}\) and \(D_{\mathrm{fake},i}\) denote these two predictions. Their normalized difference gives a gradient estimate \(g_i\) for the chunk:
\begin{equation}
    g_i=\frac{D_{\mathrm{fake},i}-D_{\mathrm{real},i}}
        {\operatorname{mean}|x_i-D_{\mathrm{real},i}|}.
    \label{eq:rmd_gradient}
\end{equation}
To propagate \(g_i\) to the generator parameters, we follow checkpointed self-forcing~\citep{savva2026solaris}. The causal rollout is generated without retaining its computation graph. A differentiable replay then recomputes each supervised prediction from its sampled noisy input and generated history, allowing \(g_i\) to be backpropagated through the generator.

The first latent frame has different statistics from later latents. Once it leaves the sliding window, the generator must predict without that distinctive frame, a condition absent from short training rollouts that can cause flicker. Following Self-Forcing~\citep{huang2025selfforcing}, we mask attention to the initial block when replaying predictions in the latter half of the training window.

\subsection{Refinement with Video-Level DMD}
\label{sec:refinement}

Matching chunk marginals improves their appearance but does not directly constrain the transitions between them. Individually realistic chunks can still form a video with inconsistent appearance or motion. We therefore refine the AR generator with the video-level DMD objective in Equation~\ref{eq:video_dmd_objective}.

The generator still produces causal self-rollouts, but the video teacher and online fake-score network now evaluate each complete rollout jointly. Their score difference updates the generator through the same differentiable replay, supervising motion and consistency across chunks. This stage begins from a generator already trained using the rollout-marginal objective. We use smaller learning rates to limit the changes made during refinement.

\section{Experiments}
\label{sec:experiments}

\subsection{Experimental Setup}
\label{sec:exp_setup}

\paragraph{Implementation Details.}
We distill the pretrained Wan 14B video diffusion model into a causal Wan 1.3B generator, initialized from Causal Forcing's causal ODE checkpoints~\citep{zhu2026causal}. RMD is trained in two stages using a metadata set comprising 70,000 unique prompts sourced from Self Forcing~\citep{huang2025selfforcing}. Our models are trained under a strictly fixed 81-frame horizon, without performing longer rollouts during the training phase.

All trainable networks are optimized using AdamW~\citep{loshchilov2019adamw} ($\beta_1=0$, $\beta_2=0.999$, weight decay $0.01$) with a global batch size of 8. During the rollout-marginal training stage, we alternate 5 fake-score updates for every 1 generator update. When the chunk size is 1, both the generator and fake-score networks use a learning rate of $2\times10^{-5}$, and we train for 900 steps. When the chunk size is 3, the generator learning rate remains $2\times10^{-5}$, while the fake-score learning rates are reduced to $2\times10^{-7}$, and we train for 750 steps. In the subsequent video-level refinement stage, the generator and fake-score learning rates are reduced to $2\times10^{-6}$ and $4\times10^{-7}$, respectively, and both models are trained for 800 steps.

\paragraph{Evaluation Protocol.}
We adopt the VBench-Long suite~\citep{huang2023vbench,huang2025vbenchpp} for our primary evaluation, using the complete standard set of 944 unique prompts with five generations per prompt. All videos are generated at $832\times480$ resolution and 16 FPS, and evaluated at a duration of approximately 60 seconds for both chunk sizes. The 500-second rollout in Figure~\ref{fig:rmd} is a qualitative example and is not included in these aggregate metrics.

\subsection{Comparison with Baselines}
\label{sec:main_results}

\begin{figure*}[t]
    \centering
    \includegraphics[width=\linewidth]{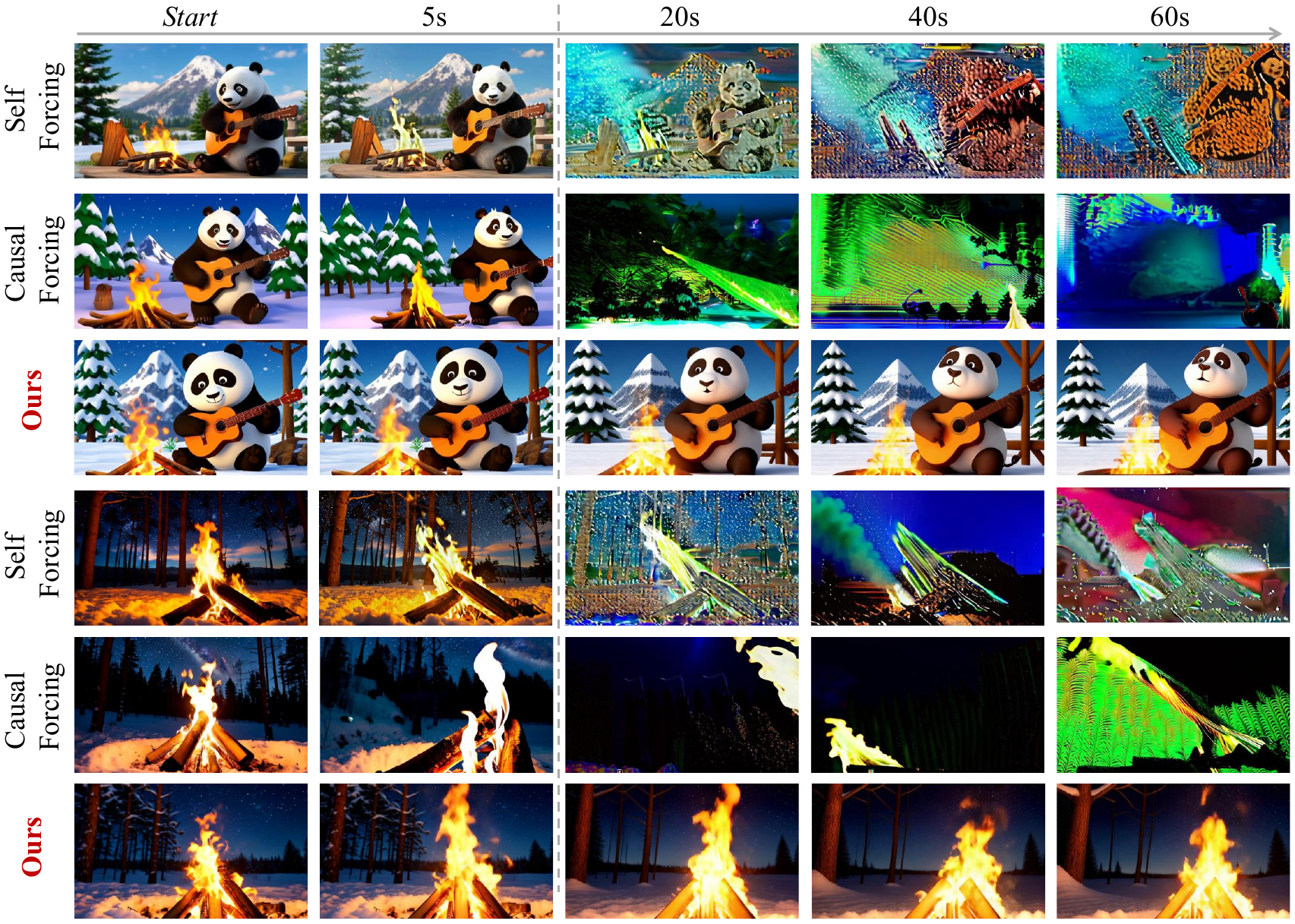}
    \caption{\textbf{Qualitative comparison of long-horizon generation.} While baseline
    methods suffer from severe color saturation and structural collapse beyond their training
    horizon, RMD consistently preserves sharp subject details and stable scene structures.}
    \label{fig:qualitative}
\end{figure*}

We compare RMD with Self Forcing~\citep{huang2025selfforcing} and Causal Forcing~\citep{zhu2026causal}. Both baselines share our standard sliding-window generation protocol, operating entirely without fixed frame anchors or modifications to the underlying autoregressive logic. This isolates the distillation objectives and ensures RMD introduces zero additional computational overhead during inference. We report results separately using chunk sizes of 1 and 3.

Figure~\ref{fig:qualitative} presents corresponding qualitative comparisons over extended rollouts. Within the 5-second training horizon, all methods generate reasonable frames. However, as generation progresses far beyond this boundary (20s, 40s, 60s), the baselines exhibit visual degradation including color saturation and structural collapse. In contrast, RMD consistently preserves clear subject details and stable scene structures throughout the 1-minute trajectory.

Table~\ref{tab:main_results} presents the quantitative comparison on 1-minute generated videos. RMD establishes a clear advantage across both configurations. With a chunk size of 1, our method improves the total score by over 10 absolute points compared with the baselines (81.26 vs. 70.94 and 69.88), while attaining quality and semantic scores of 84.52 and 68.24, respectively. With a chunk size of 3, RMD achieves a total score of 81.48, with corresponding quality and semantic scores of 85.03 and 67.24. Together, these quantitative and qualitative results demonstrate that explicitly decoupling chunk supervision from historical context effectively prevents the degradation typical of joint video-level distillation.

\begin{table}[t]
\centering
\caption{\textbf{Quantitative comparison on long-horizon generation.} Evaluated on
VBench-Long, RMD consistently outperforms video-level distillation baselines across all metrics.}
\label{tab:main_results}
\small
\setlength{\tabcolsep}{10pt}
\begin{tabular}{clccc}
\toprule
Chunk size & Method & Total $\uparrow$ & Quality $\uparrow$ & Semantic $\uparrow$ \\
\midrule
 & Self Forcing & 70.94 & 77.89 & 43.17 \\
1 & Causal Forcing & 69.88 & 78.34 & 36.03 \\
 & RMD (ours) & \textbf{81.26} & \textbf{84.52} & \textbf{68.24} \\
\midrule
 & Self Forcing & 77.44 & 81.99 & 59.22 \\
3 & Causal Forcing & 74.88 & 80.53 & 52.27 \\
 & RMD (ours) & \textbf{81.48} & \textbf{85.03} & \textbf{67.24} \\
\bottomrule
\end{tabular}
\end{table}

\subsection{Long-Horizon Extrapolation}
\label{sec:extrapolation}

\begin{figure}[t]
    \centering
    \includegraphics[width=\linewidth]{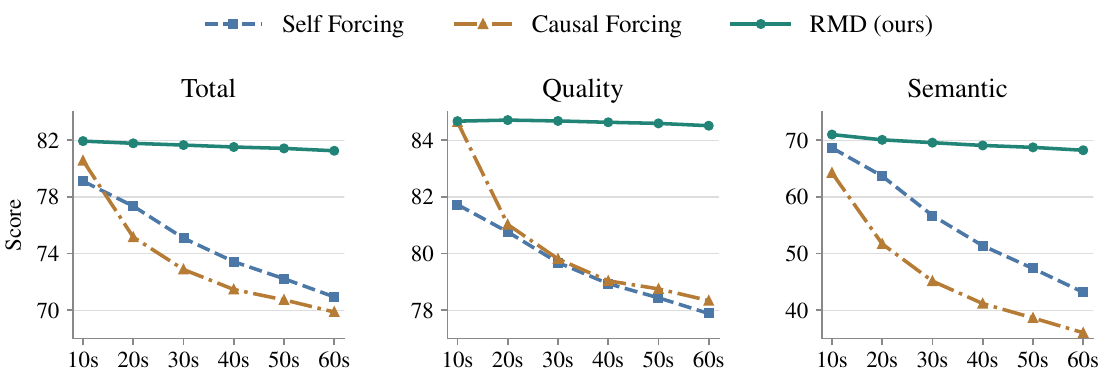}
    \caption{\textbf{Performance over increasing rollout lengths.} VBench-Long scores are
    evaluated at durations ranging from 10 to 60 seconds. As generation extends, RMD
    maintains stable visual quality and significantly mitigates the performance degradation
    suffered by baseline methods.}
    \label{fig:horizon_curve}
\end{figure}

To assess performance degradation over extended rollouts, we evaluate the same videos generated with a chunk size of 1 at durations ranging from 10 to 60 seconds. Figure~\ref{fig:horizon_curve} plots the performance dynamics over these trajectories. RMD maintains stable visual quality as the rollout extends: its quality score changes only from 84.68 at 10 seconds to 84.52 at 60 seconds, where it retains a semantic score of 68.24. In contrast, the baselines degrade steeply as prediction errors compound; for example, the total score of Self Forcing drops from 79.12 to 70.94 over the same interval. This robust extrapolation confirms that independent chunk supervision effectively mitigates the accumulation of visual artifacts.

\subsection{Ablation Studies}
\label{sec:ablation}

\begin{figure*}[t]
    \centering
    \includegraphics[width=\linewidth]{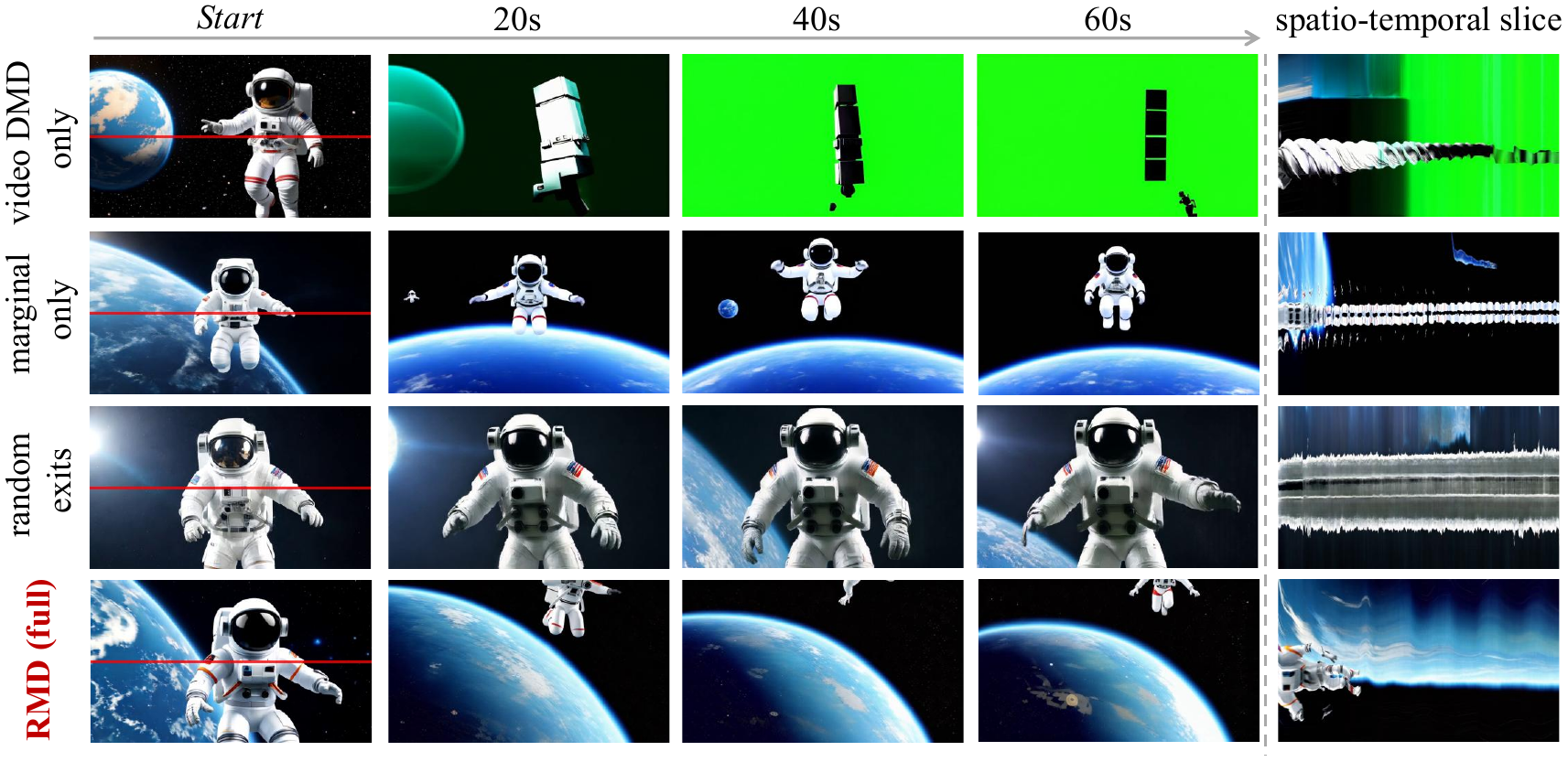}
    \caption{\textbf{Qualitative ablation results.} The spatio-temporal slice is extracted along the
    red line. Removing the rollout-marginal objective (\textit{video DMD only}) causes severe
    color artifacts. Removing video-level refinement (\textit{marginal only}) introduces
    temporal flickering (jagged slice artifacts). Removing asymmetric denoising exits
    (\textit{random exits}) destroys the temporal prior, resulting in severe temporal jitter.
    Full \textit{RMD} achieves both high visual quality and temporal coherence.}
    \label{fig:ablation_qualitative}
\end{figure*}

\paragraph{Effectiveness of the Rollout-Marginal Objective.}
Table~\ref{tab:stage_ablation} evaluates the components of RMD using a chunk size of 1. Applying video-level DMD alone yields a total score of 72.44, and qualitatively, the generation rapidly collapses into severe color artifacts (Figure~\ref{fig:ablation_qualitative}). Optimizing exclusively for the rollout-marginal objective (\textit{Marginal only}) raises the quality score to 84.90, confirming that context-free scoring effectively prevents the visual degradation.

\paragraph{Effectiveness of Refinement with Video-Level DMD.}
While the \textit{Marginal only} variant achieves high visual realism, it lacks cross-chunk constraints and falls short on temporal coherence, which is visibly manifested as jagged flickering in its spatio-temporal slice (Figure~\ref{fig:ablation_qualitative}). The full RMD pipeline resolves this via subsequent refinement with video-level DMD, which improves subject consistency, background consistency, and motion smoothness (Table~\ref{tab:stage_ablation}(b)). As shown in Figure~\ref{fig:ablation_qualitative}, this combined approach achieves the best total and semantic scores (81.26 and 68.24, respectively) while restoring smoother, more continuous trajectories in the spatio-temporal slice.

\paragraph{Effectiveness of Asymmetric Denoising Exits.}
A variant that applies random-step exits uniformly across all chunks suffers from severe temporal jitter (Figure~\ref{fig:ablation_qualitative}) and performs worse across all metrics, with total and quality scores of 78.29 and 82.48, respectively (Table~\ref{tab:stage_ablation}(a)). This supports our asymmetric design: supervising the initial chunk across random steps preserves the pretrained temporal prior, while restricting continuation chunks to the final exit strictly refines local details.

\begin{table}[t]
\centering
\caption{\textbf{Quantitative ablation results.} (a) Removing refinement with video DMD
(\textit{marginal only}) or asymmetric exits (\textit{random exits}) degrades overall
performance. (b) Temporal metrics confirm that this refinement is essential for temporal
coherence.}
\label{tab:stage_ablation}
\footnotesize
\setlength{\tabcolsep}{3pt}
\begin{minipage}[t]{0.51\linewidth}
\centering
\textbf{(a) Aggregate scores}\par\smallskip
\begin{tabular}{lccc}
\toprule
Variant & Total $\uparrow$ & Quality $\uparrow$ & Semantic $\uparrow$ \\
\midrule
video DMD only & 72.44 & 78.68 & 47.50 \\
marginal only & 80.64 & \textbf{84.90} & 63.59 \\
random exits & 78.29 & 82.48 & 61.56 \\
\midrule
RMD (full) & \textbf{81.26} & 84.52 & \textbf{68.24} \\
\bottomrule
\end{tabular}
\end{minipage}\hfill \begin{minipage}[t]{0.47\linewidth}
\centering
\setlength{\tabcolsep}{2pt}
\textbf{(b) Temporal metrics}\par\smallskip
\begin{tabular}{lrr}
\toprule
Metric & Marginal only & RMD (full) \\
\midrule
Subject Cons. $\uparrow$ & 97.29 & \textbf{98.05} \\
Background Cons. $\uparrow$ & 96.26 & \textbf{96.89} \\
Temporal Flickering $\uparrow$ & 99.06 & \textbf{99.41} \\
Motion Smoothness $\uparrow$ & 98.79 & \textbf{98.96} \\
\bottomrule
\end{tabular}
\end{minipage}
\end{table}

\section{Conclusion}
Autoregressive video generation inherently suffers from error accumulation because joint temporal scoring forces the teacher to compromise local visual quality for sequence-level consistency. To resolve this coupling, we introduce Rollout-Marginal Distillation (RMD), which disentangles spatial realism from temporal coherence. By independently matching chunk marginals and subsequently refining global dynamics, RMD provides a clean, context-free supervision signal. Quantitative evaluations over 60-second rollouts demonstrate that RMD effectively prevents artifact accumulation beyond the training horizon, while a 500-second qualitative example illustrates its potential at substantially longer durations. We believe this decoupled paradigm offers a robust foundation for future long-horizon, streamable video generation.

\bibliographystyle{iclr2027_conference}
\bibliography{iclr2027_conference}

\end{document}